\documentclass[runningheads]{llncs}
\usepackage{graphicx}
\usepackage{multirow}
\usepackage{booktabs}
\usepackage{url}
\usepackage{color}
\newcommand{\etal}{\textit{et al.}~}

\begin{document}
\title{Are pathologist-defined labels reproducible? Comparison of the TUPAC16 mitotic figure dataset with an alternative set of labels}
\titlerunning{Comparison of TUPAC16 Labels with an Alternative Set}
%
\author{Christof A. Bertram \inst{1} \and Mitko Veta \inst{2} \and Christian Marzahl \inst{3} \and Nikolas Stathonikos \inst{4} \and Andreas Maier \inst{3} \and Robert Klopfleisch \inst{1} \and Marc Aubreville \inst{3}}
\authorrunning{Bertram et al.}
%
\institute{Institute of Veterinary Pathology, Freie Universität Berlin, Berlin, Germany \and Medical Image Analysis Group, Eindhoven University of Technology, Eindhoven, Netherlands \and Pattern Recognition Lab, Computer Science, Friedrich-Alexander-Universität Erlangen-Nürnberg, Erlangen, Germany \and Department of Pathology, University Medical Center Utrecht, Utrecht, Netherlands
\email{marc.aubreville@fau.de}\\
}
\maketitle              
\begin{abstract}
Pathologist-defined labels are the gold standard for histopathological data sets, regardless of well-known limitations in consistency for some tasks. To date, some datasets on mitotic figures are available and were used for development of promising deep learning-based algorithms. In order to assess robustness of those algorithms and reproducibility of their methods it is necessary to test on several independent datasets. The influence of different labeling methods of these available datasets is currently unknown. 
To tackle this, we present an alternative set of labels for the images of the auxiliary mitosis dataset of the TUPAC16 challenge. Additional to manual mitotic figure screening, we used a novel, algorithm-aided labeling process, that allowed to minimize the risk of missing rare mitotic figures in the images. All potential mitotic figures were independently assessed by two pathologists. 
The novel, publicly available set of labels contains 1,999 mitotic figures (+28.80\%) and additionally includes 10,483 labels of cells with high similarities to mitotic figures (hard examples). 
We found significant difference comparing $F_1$ scores between the original label set (0.549) and the new alternative label set (0.735) using a standard deep learning object detection architecture. The models trained on the alternative set showed higher overall confidence values, suggesting a higher overall label consistency.
Findings of the present study show that pathologists-defined labels may vary significantly resulting in notable difference in the model performance. Comparison of deep learning-based algorithms between independent datasets with different labeling methods should be done with caution. 
\keywords{breast cancer, mitotic figures, computer-aided annotation, deep learning}
\end{abstract}

\section{Introduction}

Deep learning-based methods have shown to be powerful in the development of automated image analysis software in digital pathology. This innovative field of research has been fostered by creation of publicly available data sets of specific histological structures. One of the most extensively researched cell structures in current literature are mitotic figures (microscopic appearance of a cell undergoing cell division) in neoplastic tissue. Quantification of the highest density of mitotic figures is one of the most important histological criteria for assessment of biological tumor behavior and this pattern has therefore drawn much research attention for computerized methods. 

Manual enumeration of mitotic figures by pathology experts has some limitations including high inter-rater inconsistency of pathologists in classifying individual cells as mitotic figures as they exhibit a high degree of morphological variability and similarity to some non-mitotic structures. In previous studies, disagreement of classification occurred in 6.4-35.3\% \cite{malon2012mitotic}, and 68.2\% \cite{veta2015assessment} of labels. This calls for algorithm-assisted approaches in order to increase reproducibility as it has been proven that algorithms can have substantial agreement with pathologists on the object level \cite{veta2016mitosis}. Poor consistency of expert classification is, however, also a potential bias for deep learning-based methods, as pathologists are the current gold standard for assessment of morphological patterns, including mitotic figures, and creation of histological ground truth datasets. Due to the high inter-observer discordance of pathologists, we suspect some variability in assigned labels if images are annotated a second time. The usage of pathologist-defined labels for machine learning methods are thus somewhat a paradox as algorithmic methods, which are trained with and tested on these partially noisy ground truths, aim to overcome cognitive and visual limitations of pathologists.

In order to assess the robustness of algorithms and the reproducibility of newly developed deep learning-based methods it is necessary to test on several independent ground truth datasets. For these aspects, images should be independent but the ground truth should ideally be consistent throughout the datasets. To date, several open access datasets are available with labels for mitotic figures in digitalized microscopy images of human breast cancer \cite{Roux:2014tt,Roux:2013kn,veta2018predicting} and canine cutaneous mast cell tumors \cite{bertram2019large}, which have been developed by three research groups with somewhat variable labeling methods. As several publications have compared their algorithmic approaches between these publicly available datasets (for example \cite{aubreville2019learning,akram2018leveraging,li2019weakly}), a strong difference in test performance is known for these datasets. However, the influence of variability in the ground truth labels on training and test performance is currently unknown. In the present work, we have developed an alternative ground truth dataset for one of those publicly available images sets and assessed the difference to the original dataset. This was done using a new labeling methodology, targeted towards improved identification of mitotic figure events, and supported by the use of deep learning.

\section{Related Work}

\begin{figure}
    \centering
    \includegraphics[width=\textwidth]{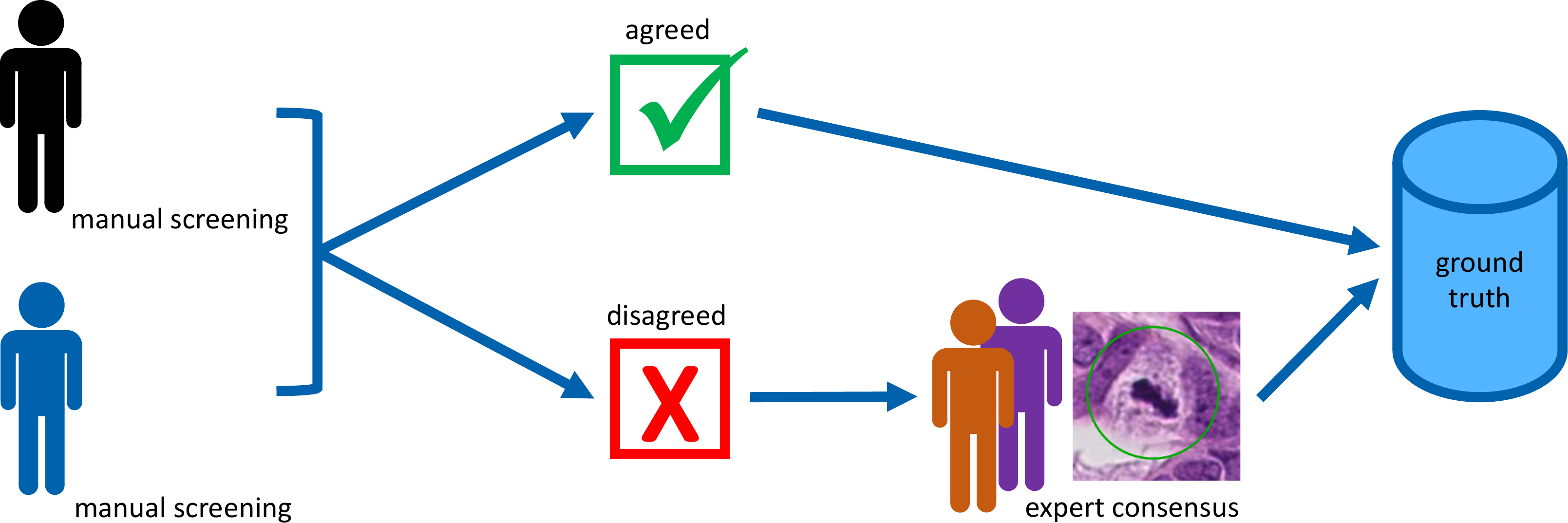}
    \caption{Annotation workflow in the original AMIDA13/TUPAC16 data sets \cite{veta2015assessment,veta2018predicting}. The images were independently screened by two pathologists. All agreed mitotic figures were directly accepted as ground truth, while disagreed cases were submitted to a panel of two additional experts.}
    \label{fig:tupac_label_method}
\end{figure}

Most publicly available data sets with annotations for mitotic figures are from human breast cancer, due to the high prevalence and high prognostic importance of the mitotic count for this tumor type. Roux \etal were the first to present a data set, consisting of five cases scanned by two whole slide scanners (and one multi-spectral scanner) and annotated by a single pathologist (ICPR MITOS 2012, \cite{Roux:2013kn}). The biggest limitation of this dataset was the potential overlap of training and test images, which were retrieved from  different locations of the same whole slide images (WSI). A year later, the MICCAI AMIDA 13 challenge introduced a new data set, covering in total 23 cases, which were evenly spread between training and test set \cite{veta2015assessment}. They were the first to acknowledge potential bias (inter-rater variance) by a single pathologist and thus perform the task by two pathologists independently, with a panel of two additional pathologists judging discordant annotations (see Fig.
~\ref{fig:tupac_label_method}). The following year, the group behind the MITOS 2012 data set introduced an extended data set at ICPR (ICPR MITOS 2014, \cite{Roux:2014tt}), consisting of 16 cases (11 for training and 5 for test), again scanned using two scanners, but this time including annotations from two pathologists. In case the pathologists disagreed, a third pathologist decided for the particular cell. The data sets includes also an expert confidence score for each mitotic figure as well as for cells probably not mitotic figures (hard negative cells). The most recent mitotic figure dataset was part of the TUPAC16 challenge \cite{veta2018predicting}, incorporating all 23 AMIDA13 cases in the training set in addition to 50 new training cases and 23 new test cases. This dataset comprises the currently the highest number of mitotic figure labels in human breast cancer.

Data about the agreement of experts in the MITOS 2014 data set can be extracted from the labels given by the challenge. Out of all 1,014 cells that were flagged by at least one pathologist as \textit{mitosis} or \textit{probably mitosis}, only 317 (31.26\%) were agreed by all pathologists to be mitotic figures, but for 749 (73.87\%) the expert consensus was \textit{mitosis}. For the MICCAI AMIDA 13 data set, Veta \etal reported an agreement in 649 out of 2038 (31.84\%) annotated cells by the two initial readers, and the consensus found 1157 (56.77\%) to be actual mitotic figures \cite{veta2015assessment}. The fact that for both data sets the final consensus strongly exceeds the initial agreement highlights that spotting of rare mitotic figure events is a difficult component in the labeling process which might lead to data set inconsistency. 

For data sets, inclusion of real-life variance of stain and tissue quality is an advantage, as the data is much more representative of a realistic use case. Current datasets on mitotic figures exhibit some differences in staining and other characteristics causing a certain domain shift \cite{aubreville2019learning} and somewhat limiting dataset transferability / robustness. Of the aforementioned datasets, the TUPAC16 dataset likely includes the highest variability due to inclusion of currently highest number of cases that were retrieved from three laboratories and scanned with two different scanners \cite{veta2018predicting}. The consequence of the higher variability is an increased difficulty for the pattern recognition task of automatic mitotic figure detection, as also reflected by lower recognition scores achieved on the data set compared to the other data sets. However, this variability represents a more realistic use-case, and is highly beneficial for the development of algorithms to be used in heterogeneous clinical environments.

\section{Material and Methods}

\subsection{Development of an Alternative Set of Labels}
\begin{figure}[tb]
    \centering
    \includegraphics[width=\textwidth]{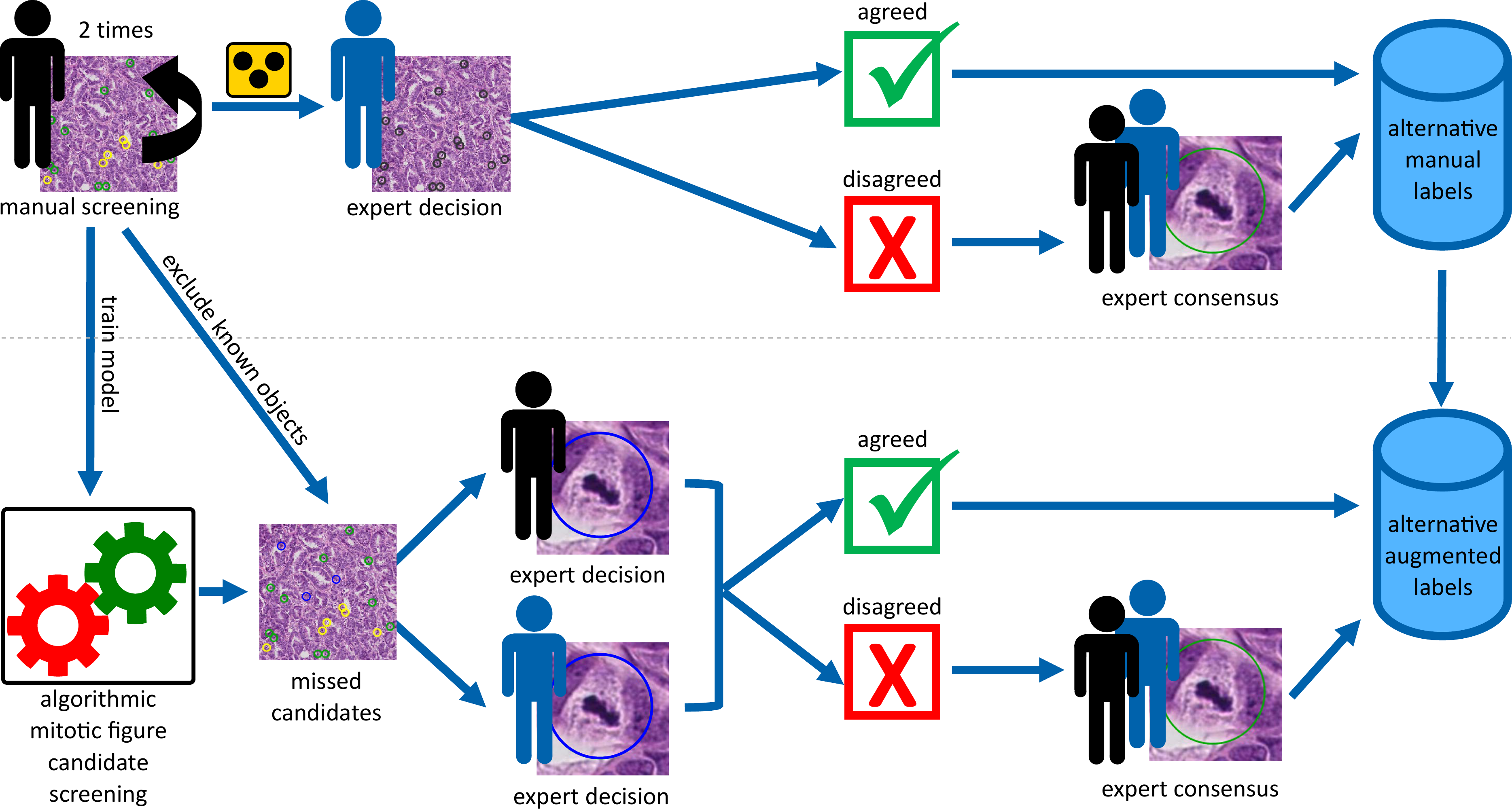}
    \caption{Labeling approach used for the creation of the alternative set of labels consisting of two steps. First, an expert screened the slides twice, and another expert performed a blind evaluation of all cells. Second, an algorithmic pipeline was used to detect cells potentially missed by the manual screening. For both steps, disagreed labels were re-evaluated by both experts in order to find a common consensus.}
    \label{fig:tupac_label_method_new}
\end{figure}

Due to the relevance of the TUPAC16 dataset (see above), we have decided to use these images in the present study for assessment of reproducibility of pathologist-defined labels. Available images from the TUPAC16 test and training set (N = 107 cases \cite{veta2018predicting}) were retrieved from the TUPAC challenge website. Cases from the AMIDA13 challenge were available as several separate, but often flanking image sections, which we stitched to single images by utilizing correlation at the image borders, wherever possible. The alternative dataset was developed in a similar way as published by Bertram \etal \cite{bertram2019large}: First, one pathology expert screened all images twice (see Fig. \ref{fig:tupac_label_method_new}) with an open source software solution with a guided screening mode \cite{aubreville2018sliderunner}. Mitotic figures (MF) and similar structures (hard negatives, HN) were labeled. The dataset from the first screening of the training set included 5,833 labels (2,188 MF; 3,645 HN), and from the second screeing 7,220 labels (2,218 MF; 5,002 HN). 

The dataset was given to a second pathologist, who assigned a second label (MF or HN) in a blinded manner (label class obscured) supported by the annotation software through automatic presentation of image section with unclassifed objects. The second pathologist assigned the MF label in 2,272 cases and the HN label in 4,978 cases. Initial agreement for the class MF was found for 1,713 cells (61.69\%), the pathologists disagreed on 1,064 cells (14.74\% of all cells). All disagreed cells were re-evaluated by both experts, and the consensus of the manual dataset contained 1,898 MF and 5,340 HN.

Subsequently, labels from the first expert were used for an algorithmic-aided pipeline for detection of missed objects, like described in \cite{bertram2019large}. The pipeline extracted image patches around additionally detected mitotic figure candidates, sorted according to their model score. The algorithm-based screening additionally found 5,824 objects (mitotic figure candidates), which were then extracted as 128x128\,px image patches centered around the detection. Two experts assessed (MF or HN) these patches independently and agreed on all but 142 patches. All agreed objects were assigned to the dataset immediately and disagreed objects were re-evaluated by joint assessment for consensus. The final augmented data set contains 1,999 MF and 10,483 HN. Please note that all numbers are given only for the training part of the set to not reveal information about the test set for further usage.

\subsection{Automatic Mitosis Detection Methods}
We evaluated the alternative labels using a standard, state-of-the-art object detection approach: We customized RetinaNet \cite{lin2017focal} based on a pre-trained ResNet-18 stem with an input size of $512\times512$\,px to have the object detection head only attached at the $32\times32$ resolution of the feature pyramid network. We chose four different sizes (scales) of bounding boxes to enable augmentation by zooming and rotation, but only used an 1:1 aspect ratio, since the bounding boxes were defined to be squares. We randomly chose 10 tumor cases to be our validation set, which was used for model selection based on the mAP metric. After model selection, we determined the optimum detection threshold on the concatenated training and validation set. Models were trained on both, the original TUPAC16 label set and the novel, alternative set, and evaluated on the respective test sets using $F_1$ as metric.

Additionally, we calculated the model scores for individual cells of the data sets to assess model confidence. Since the test set of both label sets are not available publicly, we used a three-fold cross-validation on the training set. For this, we disabled the threshold normally used within the non-maximum-suppression of the model post-processing, which enabled us to derive model scores from the classification head of the model for all cells of our data set. We matched annotations in both data sets under the assumption that that all annotations within a distance of 25 pixels refer to the same object.

The complete training set and all code that was used for the evaluation is made available online\footnote{\url{https://github.com/DeepPathology/TUPAC_alternativeLabels}}. We encourage other research groups to use this alternative dataset for training their algorithms and we will provide evaluation of the performance of detection results on the test set upon a reasonable request to the corresponding author. 


\section{Results}

\begin{figure}[t]
    \centering
    \includegraphics[width=\textwidth]{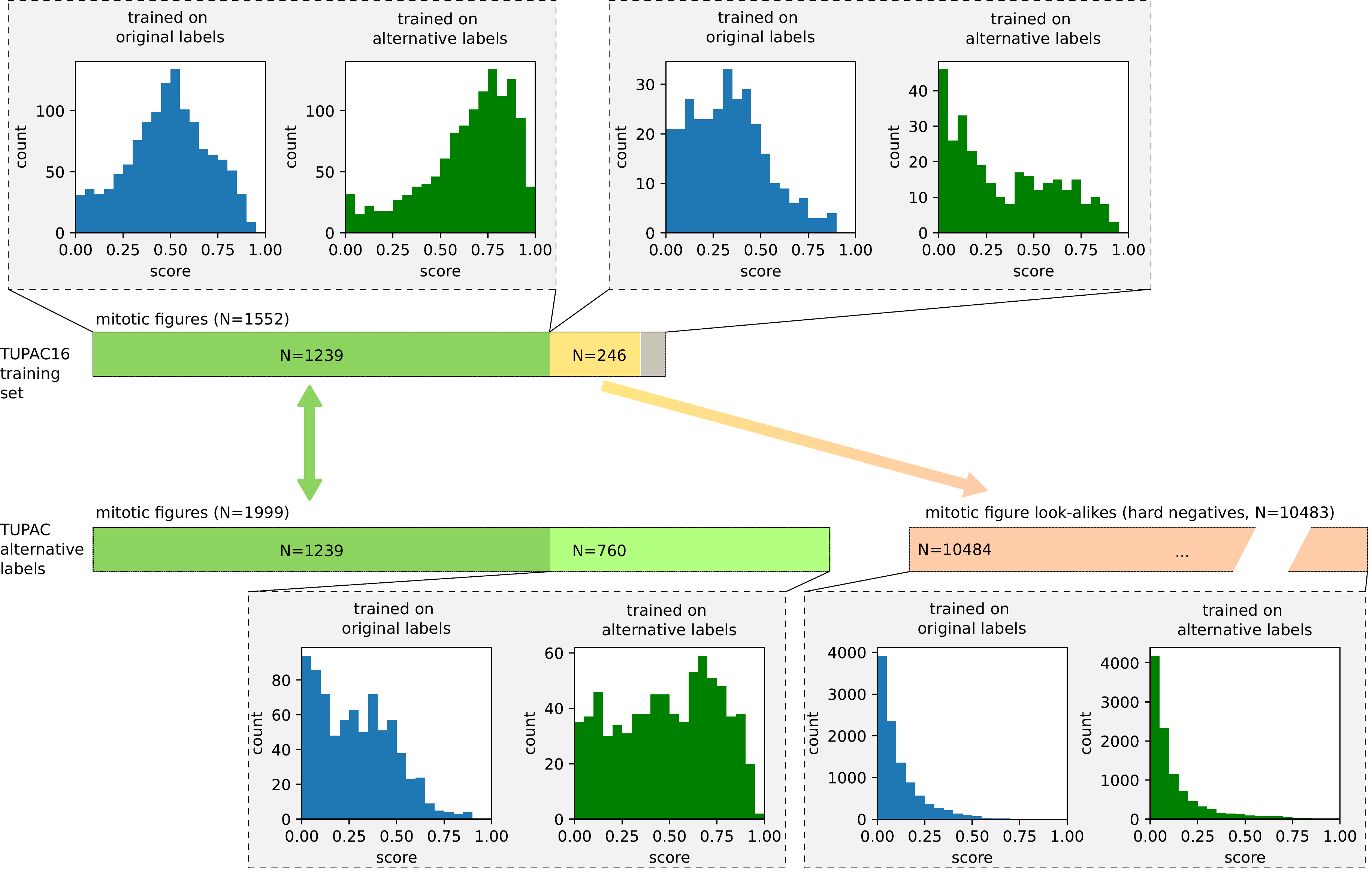}
    \caption{Comparison of the original TUPAC16 and alternative label sets (training part only). The two expert teams agreed upon 1,239 mitotic figures, while the new set contains 760 additional labels for mitotic figures and 246 out 309 disagreed cells were labeled as hard negatives. The plot shows also the concatenated model scores given by a RetinaNet-approach trained in three-fold cross-validation on the original (blue) and alternative (green) label set. }
    \label{fig:results_distribution}
\end{figure}

Comparing the original and the new, alternative training label sets, we find that they agree for 1,239 MF annotations (53.59\%), while the two expert groups disagreed on 1,073 cells (46.41\%). As depicted in Fig. \ref{fig:results_distribution}, 246 of MF identified in the original TUPAC16 label set were assigned to be hard examples in the alternative set, while 67 were not annotated at all. The alternative set assigned 760 further cells with the label MF, that were not annotated in the original label set. 

Looking at the concatenated model scores from the cross-validation experiment, we can state that the model trained on the alternative set shows an overall higher confidence for agreed mitotic figures. In contrast, MF labels only present in the original TUPAC16 dataset had an overall lower model score with a tendency towards higher values in the models trained on the original set (median values are 0.326 and 0.284). The group of labels newly assigned in the alternative set shows  higher scores for the model trained on the alternative set (median value of 0.503 vs. 0.266), while the group of hard negatives has a very similar distribution with low model scores for both training label sets.

\begin{table}[]
\setlength{\tabcolsep}{6pt}
    \resizebox{\linewidth}{!}{
    \centering
    \begin{tabular}{c|c|ccc}
         \multirow{2}{*}{metric} & \multirow{2}{*}{training} & \multicolumn{3}{c}{test} \\
        \cline{3-5}
        & & TUPAC original & alternative(manual)  & alternative(augmented) \\
        \hline
        \multirow{2}{*}{\textbf{$F_1$ score}} & TUPAC original & 0.549 & 0.587 & 0.563 \\
        \cline{2-5}
         & alternative (augmented) & 0.555 & 0.719 & 0.735 \\
        \hline
        \multicolumn{5}{c}{} \\
        \hline
        \multirow{2}{*}{\textbf{precision}} & TUPAC original & 0.540 & 0.682 & 0.699 \\
        \cline{2-5}
         & alternative (augmented) & 0.477 & 0.713 & 0.772 \\
        \hline
        \multicolumn{5}{c}{} \\
        \hline
        \multirow{2}{*}{\textbf{recall}} & TUPAC original & 0.559 & 0.515 & 0.471 \\
        \cline{2-5}
         & alternative (augmented) & 0.665 & 0.725 & 0.701 \\
        \hline
    \end{tabular}
    }
    \caption{Comparison of $F_1$ score, precision and recall achieved on the different label sets with a customized RetinaNet\cite{lin2017focal} approach.}
    \label{tab:results_comparison}
\end{table}

The higher model confidence for mitotic figures on the alternative dataset in Fig. \ref{fig:results_distribution} coincides with a generally higher $F_1$ score in model performance on the test set (see Table \ref{tab:results_comparison}). We see a small increment for using the data set using the machine-learning-aided detections for potentially missed cells, related to a notable increase in precision.

\section{Discussion}
Labeled data is the foundation for training and testing of deep learning-based algorithms. Although a vast diversity of labeling methods have been applied for mitotic figure dataset development \cite{bertram2019large,marzahl2020fast,Roux:2013kn,Roux:2014tt,veta2015assessment,veta2018predicting}, the effects of these methods on algorithmic performance is currently not fully understood. With recent improvements of deep learning methods, the demand for high-quality data is also increasing. The currently highest reported $F_1$ score on the original TUPAC16 dataset is 0.669 \cite{li2019weakly}, which is significantly higher than the value achieved by our standard RetinaNet approach on the same labels ($F_1$ score: 0.549). Considering the difference between the present and the state-of-the-art results by Li \etal \cite{li2019weakly} on the original TUPAC16 dataset, it seems likely that also the results on the alternative datasets may be further improved by more advanced methods, which we encourage as we have made the alternative datasets publicly available. However, instead of aiming to achieve highest possible performance, we wanted to assess effects of using different ground truth datasets of the same images with the same deep learning method. The major finding of the present study was that pathologists-defined labels are not necessarily reproducible even when using annotation protocols that take the consensus of several experts as the ground truth, and differences may lead to notable variation in performance. In this case the model trained and tested on the alternative dataset yielded an higher $F_1$ score of 18.6 percentage points compared to the same model architecture trained and tested on the original label set. The present results indicate that comparing model performance between two different datasets should be done with caution.

The alternative datasets contains 28.80\% more mitotic figure labels in the training set. Some of these additional mitotic figures have a relatively low model score, which could question the unambiguous nature of the labels regardless of the overall higher $F_1$ score. However, the increased model scores for the algorithms trained on the alternative data, in comparison to the original data, indicates a overall higher consistency. Regardless, both datasets include numerous labels with low model score, which could potentially be explained by the high morphological variability of mitotic figures and availability of very few patches of some morphological variants for training. Large-scale datasets with even higher numbers of mitotic figure labels might potentially overcome this limitation. Additionally, different degrees of inconsistency have been described between pathologists  \cite{malon2012mitotic,Roux:2014tt,veta2018predicting} and pathologists-defined labels represent a somewhat noisy reference standard regardless agreement or consensus by several pathologists. 

Besides the difficulties in the classification of mitotic figures, differences of expert-defined labels may arise from lack of identifying rare events \cite{veta2016mitosis}. The higher number of mitotic figure labels with presumably high label consistency in the alternative datasets (see above), suggests that fewer mitotic figures were overlooked. The labeling method of the alternative dataset basically follows the paradigm of Viola and Jones \cite{viola2001rapid}, of having an initially highly sensitive detection, followed by a secondary classification with high specificity achieved through dual expert consensus. High sensitivity in detecting potential mitotic figure labels was achieved by repeated manual screening of the images and an additional algorithmic augmentation. As the algorithmic detection of missed objects may potentially introduce a confirmation bias, image patches were reviewed by two pathologists independently. Final agreement on mitotic figures was only obtained for 2.4\% of the augmented cells, illustrating the desired high sensitivity / low specificity of this approach for algorithmic mitotic figure detection. Thereby identification of presumably almost all mitotic figures (high number of true positives and low number of false negatives) was ensured. Of note, adding this relatively low number of labels to the ground truth had a notable effect on performance of up to 1.6 percentage points, consistent with previous findings \cite{bertram2019large}.

Algorithmic  approaches  for  dataset  development  have  become  more  popular  in  recent  years  due  to  increasing  demand  on  datasets that are difficult to accomplish with solely manual approaches. As described above, algorithmically supported identification of missed candidates may improve dataset quality and requires algorithms with high sensitivity \cite{bertram2019large}. In contrast, enlargement of datasets (higher quantity) may be facilitated through algorithmic detections with high specificity in order to ensured that mainly true positives and only few false positive labels are generated. This approach can be used for the creation of datasets with reduced expenditure of expert labor (crowd sourcing \cite{albarqouni2016aggnet} or expert-algorithm-collaboration \cite{marzahl2020fast}), or fully automated generation of additional data without pathologists-defined labels (pseudo-labels) \cite{akram2018leveraging}. Tellez \etal \cite{tellez2018whole} recently investigated another approach, that used an specific staining for mitotic figures (immunohistochemistry with antibodies against phosphohistone H3) with computerized detection of reference labels and subsequent registration to images of the same tissue section with standard, non-specific hematoxylin and eosin stain. Besides requiring minimal manual annotation effort, this methods may eliminate expert-related inconsistency and inaccuracy.

In conclusion, this study shows considerable variability in pathologists-defined labels. A subsequent effect was evident on training the models (variation of model scores) and performance testing (variation of $F_1$ score). This needs to be considered when robustness of algorithms or reproducibility of developed deep learning methods are to be tested on independent ground truth datasets with different labeling methods. Therefore, scores should be interpreted in relation to reference results on that specific datset. Further studies on reduction of expert-related inconsistency and inaccuracy are encouraged.


\bibliographystyle{splncs04}


\begin{thebibliography}{10}
\providecommand{\url}[1]{\texttt{#1}}
\providecommand{\urlprefix}{URL }
\providecommand{\doi}[1]{https://doi.org/#1}

\bibitem{akram2018leveraging}
Akram, S.U., Qaiser, T., Graham, S., Kannala, J., Heikkil{\"a}, J., Rajpoot,
  N.: Leveraging unlabeled whole-slide-images for mitosis detection. In:
  Computational Pathology and Ophthalmic Medical Image Analysis, pp. 69--77.
  Springer (2018)

\bibitem{albarqouni2016aggnet}
Albarqouni, S., Baur, C., Achilles, F., Belagiannis, V., Demirci, S., Navab,
  N.: Aggnet: deep learning from crowds for mitosis detection in breast cancer
  histology images. IEEE Trans Med Imag  \textbf{35}(5),  1313--1321 (2016)

\bibitem{aubreville2018sliderunner}
Aubreville, M., Bertram, C., Klopfleisch, R., Maier, A.: Sliderunner. In:
  Bildverarbeitung f{\"u}r die Medizin 2018, pp. 309--314. Springer (2018)

\bibitem{aubreville2019learning}
Aubreville, M., Bertram, C.A., Jabari, S., Marzahl, C., Klopfleisch, R., Maier,
  A.: Learning new tricks from old dogs-inter-species, inter-tissue domain
  adaptation for mitotic figure assessment. In: Bildverarbeitung f{\"u}r die
  Medizin 2020. pp.~1--7 (2020)

\bibitem{bertram2019large}
Bertram, C.A., Aubreville, M., Marzahl, C., Maier, A., Klopfleisch, R.: A
  large-scale dataset for mitotic figure assessment on whole slide images of
  canine cutaneous mast cell tumor. Scientific Data  \textbf{6}(274), ~1--9
  (2019)

\bibitem{li2019weakly}
Li, C., Wang, X., Liu, W., Latecki, L.J., Wang, B., Huang, J.: Weakly
  supervised mitosis detection in breast histopathology images using concentric
  loss. Med Imag Anal  \textbf{53},  165--178 (2019)

\bibitem{lin2017focal}
Lin, T.Y., Goyal, P., Girshick, R., He, K., Doll{\'a}r, P.: Focal loss for
  dense object detection. In: Proceedings of the IEEE international conference
  on computer vision. pp. 2980--2988 (2017)

\bibitem{malon2012mitotic}
Malon, C., {et al.}: Mitotic figure recognition: Agreement among pathologists
  and computerized detector. Anal Cell Pathol  \textbf{35}(2),  97--100 (2012)

\bibitem{marzahl2020fast}
Marzahl, C., {et al.}: Are fast labeling methods reliable? {A} case study of
  computer-aided expert annotations on microscopy slides. preprint,
  arXiv:2004.05838  (2020)

\bibitem{Roux:2014tt}
Roux, L., {et al.}: {MITOS {\&} ATYPIA - Detection of Mitosis and Evaluation of
  Nuclear Atypia Score in Breast Cancer Histological Images}. Image Pervasive
  Access Lab IPAL, Agency Sci., Technol. Res. Inst. Infocom Res., Singapore
  (2014)

\bibitem{Roux:2013kn}
Roux, L., {et al.}: {Mitosis detection in breast cancer histological images An
  ICPR 2012 contest}. Journal of Pathology Informatics  \textbf{4}(1), ~8
  (2013)

\bibitem{tellez2018whole}
Tellez, D., {et al.}: Whole-slide mitosis detection in {H\&E} breast histology
  using {PHH3} as a reference to train distilled stain-invariant convolutional
  networks. IEEE Trans Med Imag  \textbf{37}(9),  2126--2136 (2018)

\bibitem{veta2015assessment}
Veta, M., {et al.}: Assessment of algorithms for mitosis detection in breast
  cancer histopathology images. Med Imag Anal  \textbf{20}(1),  237--248 (2015)

\bibitem{veta2018predicting}
Veta, M., {et al.}: {Predicting breast tumor proliferation from whole-slide
  images - The TUPAC16 challenge}. Med Imag Anal  \textbf{54},  111--121 (May
  2019)

\bibitem{veta2016mitosis}
Veta, M., Van~Diest, P.J., Jiwa, M., Al-Janabi, S., Pluim, J.P.: Mitosis
  counting in breast cancer: Object-level interobserver agreement and
  comparison to an automatic method. PloS one  \textbf{11}(8),  e0161286 (2016)

\bibitem{viola2001rapid}
Viola, P., Jones, M.: Rapid object detection using a boosted cascade of simple
  features. In: Proceedings of the 2001 IEEE computer society conference on
  computer vision and pattern recognition. vol.~1, pp.~I--I. IEEE (2001)

\end{thebibliography}

\end{document}